\documentclass[10pt,reqno]{amsart}
\usepackage{float}
\usepackage{companypaper}
\usepackage[numbers,sort&compress]{natbib}
\usepackage{placeins}
\usepackage{nicefrac}
\CompanyLogo{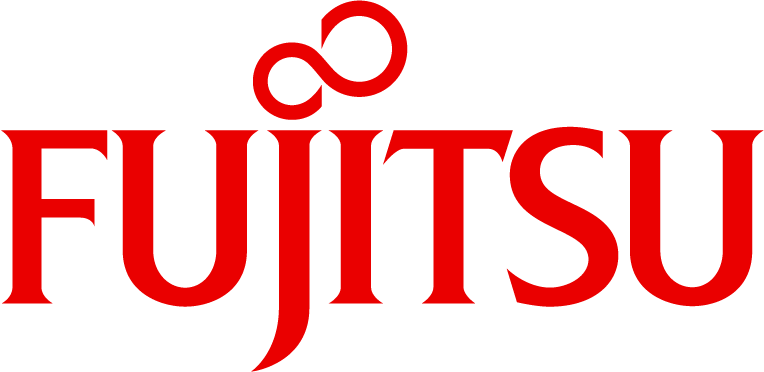}
\CompanyLogoHeight{11mm}
\CompanyHeaderText{Fujitsu Limited}
\CompanyDocType{Technical Report}
\CompanyClassification{}
\CompanyRunningTitle{Decision Model Factory}
\CompanyAffiliations{\textsuperscript{1}Fujitsu Limited,~
  \textsuperscript{2}RIKEN Center for AIP}
\CompanyCorrespondence{Yuma Ichikawa at
  \href{mailto:ichikawa.yuma@fujitsu.com}{\texttt{ichikawa.yuma@fujitsu.com}}}
\title{Decision Model Factory:\\
Building Specialized Sovereign Decision Models}
\author{Yuma Ichikawa\textsuperscript{1,2}}
\date{October 24, 2026}
\hypersetup{pdftitle={Decision Factory: Building Specialized Sovereign Decision Models},
  pdfauthor={Yuma Ichikawa},
  pdfsubject={An automated factory for specialized sovereign decision models}}

\begin{document}
\maketitle
\begin{abstract}
    Decision needs vary with each application, language, and deployment environment. The Decision Model Factory turns those needs into specialized, sovereign models on demand. A brief recipe guides the training, calibration, verification, and evaluation of a selected backbone, delivering a deployable decision model under the operator's control. The same workflow creates a new specialist whenever the requirements change. Across Japanese decision tasks, compact deployment, and long-context passage ranking, five factory-built models improve upon their backbones; three surpass the strongest results in a fixed public Japanese benchmark comparison. A dedicated runner computes only permitted-answer probabilities and reuses shared context, more than doubling serving speed on matched multi-question workloads with essentially unchanged decision scores. Together, the factory and runner make fast, task-specific decision models practical to build and deploy whenever needed.
\end{abstract}

\begin{figure}[H]
  \centering
  \includegraphics[width=\linewidth]{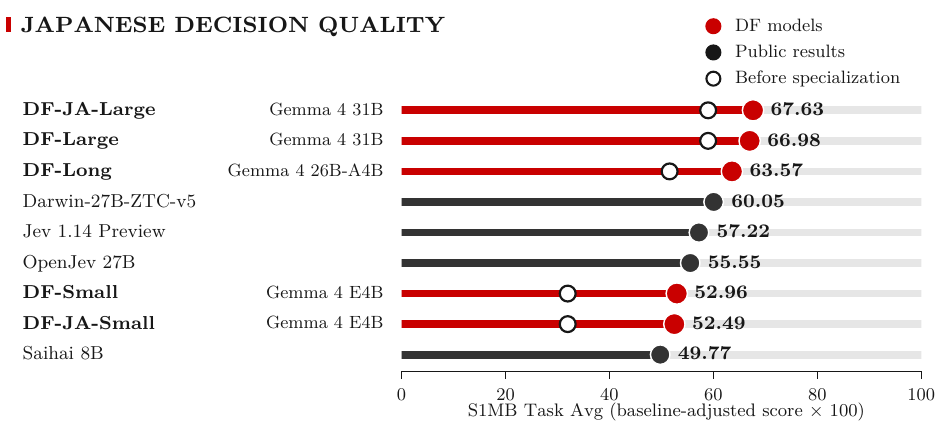}
  \caption{Japanese decision quality on the System One Mosaic Benchmark
  (S1MB). Red marks show factory-built DF models, hollow circles their
  backbones before specialization, and black marks public results
  \citep{s1mbresults2026}.}
  \label{fig:benchmarks}
\end{figure}

\section{Introduction}
Many workflows require a decision, not a paragraph: does a clause violate policy, where should a ticket be routed, or which passage supports an answer? Large language models (LLMs) can reason over the evidence, but free-form generation adds decoding and parsing without directly yielding calibrated probabilities for the permitted outcomes. Decision models \citep{almeida2026jev} instead return a distribution over those outcomes in one forward pass, so applications can act on confident predictions and refer uncertain cases for review.

The challenge is to make the correct decision model available for each job. Languages, policies, answer labels, document lengths, and serving budgets vary across applications. Japanese contract review and ticket routing on a central processing unit (CPU), for example, call for different specializations. Organizations may also need to retain control of their training data, model weights, and deployment environment. We use \emph{sovereign} for a model that the operator can build and serve under its own control. Repeating a separate training and deployment project for every requirement is difficult to sustain.

\begin{figure}[tb]
  \centering
  \includegraphics[width=\linewidth]{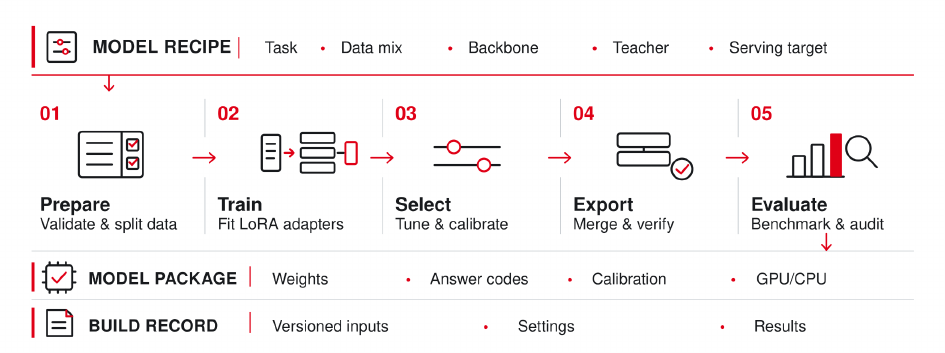}
  \caption{Decision Model Factory. A model recipe passes through five
  stages to produce a verified model package and a record of the build.}
  \label{fig:factory}
\end{figure}

Decision Model Factory makes specialization a reproducible build that can be launched whenever a new need arises. Drawing on the model-factory concept \citep{rowell2025factory}, a recipe specifies the backbone, decision task, training inputs, and serving target. The Factory runs training, calibration, export verification, and evaluation, then delivers a verified package for graphics processing unit (GPU) or CPU serving (Figure~\ref{fig:factory}). Its stages can run unattended on the operator's computing system. When a language, task, or deployment constraint changes, the recipe changes, not the pipeline.

We demonstrate this approach with five models from three Gemma~4 backbones \citep{gemmateam2026gemma4} (Table~\ref{tab:family}). Their recipes produce Japanese specialists, compact distilled variants, and a long-context model that also ranks passages. Three exceed the strongest public score in our fixed Japanese comparison. The accompanying \emph{decision runner} serves multiple judgments over shared evidence more than twice as fast as the general-purpose path in matched workloads. Together, they offer a practical way to build and deploy the next specialized decision model as needs evolve.

\begin{table}[tb]
\centering
\setlength{\tabcolsep}{4.5pt}
\caption{The DF family and S1MB Japanese Task Avg before and after
specialization. JA denotes Japanese-focused recipes; DF-Long uses about
4 billion active parameters per token.}
\label{tab:family}
    \begin{tabular}{@{}lllrr@{}}
    \toprule
    Model & Gemma 4 backbone & Recipe & Before & \cellcolor{red!6}After \\
    \midrule
    DF-JA-Large & 31B & Japanese-focused & 58.99 & \cellcolor{red!6}67.63 \\
    DF-Large & 31B & General & 58.99 & \cellcolor{red!6}66.98 \\
    DF-Long & 26B-A4B & General, long inputs & 51.58 & \cellcolor{red!6}63.57 \\
    DF-Small & E4B & General, distilled & 31.97 & \cellcolor{red!6}52.96 \\
    DF-JA-Small & E4B & Japanese-focused, distilled & 31.97 & \cellcolor{red!6}52.49 \\
    \bottomrule
    \end{tabular}
\end{table}

\section{From a Language Model to a Decision Model}
\label{sec:readout}
A decision request consists of a \emph{state}~$s$ (e.g., a ticket or document) and a \emph{question}~$q$ that specifies the task instructions and a set of $K$ possible answers. We consider three answer types: \emph{Noul} for binary yes-or-no decisions, \emph{Choice} for selecting among named alternatives, and \emph{Score} for ratings on an ordered scale with values $v_1 < \cdots < v_K$. A fixed template converts $(s,q)$ into a prompt~$x$ that presents the state first and assigns each answer option a code $c_k$ (e.g., \texttt{A} or \texttt{B}). Each code corresponds to a single token in the backbone model's vocabulary.

Let $h(x)$ denote the backbone's final hidden state at the last position of the prompt, and let $w_k$ denote the row of its output matrix corresponding to code $c_k$. We compute the logit for option $k$ as $z_k(x) = g(w_k^\top h(x))$, where $g$ represents any additional output transformation applied by the backbone. Following the option-logit readout of \citet{geuter2026}, we define the decision distribution as
\begin{equation}
  p_\theta(k\mid s,q)=\frac{\exp(\nicefrac{z_k(x)}{\tau_a})}
  {\sum_{j=1}^{K}\exp(\nicefrac{z_j(x)}{\tau_a})},~~~ k=1,\dots,K,
  \label{eq:readout}
\end{equation}
where $\theta$ denotes the model parameters, $a$ specifies the answer type, and $\tau_a>0$ represents the corresponding calibration temperature. For \emph{Score} answers, we additionally report the expected value $\sum_k p_\theta(k\mid s,q)v_k$. This option-based readout operates over the allowed answers and directly reuses the backbone's tokenizer and output layer, requiring no additional classification head.

We adapt the backbone using low-rank adaptation (LoRA) \citep{lora2021}. For each selected weight matrix $W_0\in\mathbb{R}^{m\times n}$, we keep $W_0$ frozen and learn a low-rank update $\Delta W=BA$, where $B\in\mathbb{R}^{m\times r}$, $A\in\mathbb{R}^{r\times n}$, and $r\ll\min(m,n)$. During training, we use the option distribution before temperature calibration, together with a target distribution $y$ over the allowed answers. The training loss is
\begin{equation}
  \mathcal{L}=-\sum_{k=1}^{K}y_k \log p_\theta(k\mid s,q)
  +\lambda \mathbf{1}[a=\text{Score}] \times 
  \left|\sum_{k=1}^{K}(p_\theta(k\mid s,q)-y_k )v_k\right|.
  \label{eq:loss}
\end{equation}
The indicator activates the second term only for \emph{Score} tasks, where it penalizes errors in the expected value with a weight of $\lambda\ge0$. The target distribution $y$ is either a one-hot encoding of the ground-truth answer or a distillation mixture $y=(1-\beta)y^{\mathrm{label}}+\beta p^{\mathrm{teacher}}$, where $p^{\mathrm{teacher}}$ denotes the teacher's distribution over the allowed answers and $\beta\in[0,1]$ controls its contribution \citep{hinton2015distillation}.

After training, the Factory selects the update strength
\begin{equation}
  W(\sigma)=W_0+\sigma\Delta W,~~ \sigma\in\Sigma,
  \label{eq:strength}
\end{equation}
by evaluating each candidate in $\Sigma$ on held-out development decisions. At the selected strength, the Factory then fits a separate calibration temperature $\tau_a$ for each answer type \citep{guo2017calibration}. Finally, it verifies that the exported package reproduces the selected readout within a predefined tolerance.

\section{Decision Model Factory}
\label{sec:factory}

Each build begins with a recipe specifying the backbone model, decision dataset, optional teacher model, adapter configuration, and deployment formats. The preparation stage validates the answer types and sets aside development examples. Training then learns the low-rank update using the objective in Equation~\eqref{eq:loss}, followed by the selection of the adapter strength and calibration temperatures. During export, the selected update is merged into the backbone, and the resulting model is verified to produce consistent readouts between training and serving. For mixture-of-experts (MoE) backbones, this verification is performed separately for each expert. Finally, the runner evaluates the exported model on public benchmarks. The resulting package includes the model weights, prompt template, option codes, and calibration temperatures required for deployment.

The pipeline supports unattended execution and can resume from interruptions. Each stage records the recipe, code, data hashes, and selected configurations, ensuring that evaluation results can be traced to the exact build that produced them. These results can then inform subsequent recipes, while all builds follow the same training and serving workflow.

We develop five model variants using distinct recipes. The Japanese-focused models differ in their training data mixtures and adapter ranks, while the compact models learn from a larger decision teacher through distillation. DF-Long additionally incorporates long documents and relevance judgments into its training. Training examples include demonstrations embedded in the question \citep{brown2020fewshot}, distracting content surrounding relevant evidence \citep{liu2024lost}, and instructions embedded in documents that the model must interpret as content rather than commands \citep{greshake2023injection}. Despite these differences, all five models share a common build pipeline and serving interface.

\section{A Purpose-Built Decision Runner}
\label{sec:runtime}

The General-purpose language model serves by computing logits over the entire vocabulary, even when a decision only requires a small set of allowed answers. Similarly, processing multiple questions about the same state independently can lead to redundant computations over shared evidence. Our decision runner addresses both inefficiencies.

\paragraph{Option-only output.}
Instead of projecting onto the full vocabulary, the runner projects the final hidden state $h(x)$ onto only the output-matrix rows $w_1,\dots,w_K$ corresponding to the allowed answers. It then applies the backbone's output transformation and computes the decision probabilities using Equation~\eqref{eq:readout} and the exported calibration temperatures. Consequently, the output-projection cost scales with the number of allowed answers rather than the vocabulary size, while the resulting probabilities remain equivalent to those of the full-vocabulary readout, up to floating-point rounding.

\paragraph{Shared-state execution.}
When multiple questions share the same state, the runner computes the attention keys and values for that state once and reuses the cached prefix across questions. For the MoE backbone, it further captures the prompt-processing pass in a GPU execution graph to reduce kernel-launch overhead. The CPU implementation similarly accesses only the output-matrix rows required for the allowed answers. For backbones that support reusing earlier attention states in later layers, the runner can also skip unnecessary computations at intermediate prompt positions \citep{gemmateam2026gemma4}.

\section{Experiments}
\label{sec:experiments}
\subsection{Evaluation Setting}
\label{sec:setting}
\paragraph{Models and tasks.}
We evaluate the five models listed in Table~\ref{tab:family}, with training configurations detailed in Appendix~\ref{app:settings}. The evaluated revisions of the System One Mosaic Benchmark (S1MB) \citep{s1mb2026} comprise 8,067 judgments across 27 Japanese benchmarks and 26,269 judgments across 137 English benchmarks. Each benchmark score is adjusted for its task-specific trivial baseline and clipped to $[0,1]$. We compute \emph{Task Avg} by first averaging scores within each answer type, then taking the equally weighted average across the three types and multiplying the result by 100.

DF-Long supports inputs of up to 262,144 tokens. We evaluate its passage-ranking performance on Japanese Question Answering with Retrieval Augmentation (JQaRA) \citep{tateno2024jqara} and the Japanese tracks of MIRACL \citep{zhang2022miracl} and Multilingual Long-Document Retrieval (MLDR) \citep{chen2024m3}. Candidates are ranked by the probability assigned to the affirmative option in a \emph{Noul} relevance judgment, following an approach analogous to the relevance-token scoring used by monoT5 \citep{nogueira2020monot5}. We assess robustness against embedded instructions designed to redirect the model's answer, using 300 held-out states for each of English, Japanese, and Portuguese. We also evaluate calibration on 600 development decisions using expected calibration error (ECE), defined as the weighted average of the absolute differences between confidence and accuracy across bins.

\paragraph{Comparisons.}
For S1MB, we use the official evaluator and 8-bit model packages. To isolate the effects of specialization, we evaluate each backbone before adaptation using the same readout and packaging as its corresponding DF model. Our comparison with publicly available results is based on a fixed snapshot of 34 complete evaluations retrieved on 18 October 2026 \citep{s1mbresults2026}. For passage ranking, we compare DF-Long against Ruri~v3 \citep{tsukagoshi2024ruri} and bge-reranker-v2-m3 \citep{bgereranker2024}, using the same candidate passages for all methods.

\paragraph{Serving measurements.}
We measure serving performance on a single NVIDIA B200 GPU or 16 Intel Xeon CPU cores, using freshly generated synthetic support tickets. All reported timings are medians over repeated batches after warm-up. In the runner ablation, we hold the serving engine and model weights fixed to isolate the effects of the runtime optimizations. Cross-model latency and bulk-throughput experiments use the configurations specified in their respective figures.

\subsection{Results}
\label{sec:results}
\paragraph{Decision quality.}
Specialization consistently improves the Japanese Task Avg across all backbones (Figure~\ref{fig:benchmarks}). With approximately 4 billion active parameters per token, DF-JA-Large, DF-Large, and DF-Long achieve scores of 67.63, 66.98, and 63.57, respectively, all surpassing the highest score of 60.05 in the fixed public snapshot. The distilled small models exhibit the largest improvements: DF-Small and DF-JA-Small increase their scores from 31.97 to 52.96 and 52.49, respectively, both exceeding the published score of 49.77 achieved by Saihai~8B \citep{tateno2026saihai}. Specialization also yields substantial gains on English tasks, improving DF-Long from 52.14 to 63.55 and DF-Small from 36.49 to 47.30.

\begin{table}[tb]
    \centering
    \setlength{\tabcolsep}{4pt}
    \caption{DF-Long across tasks. S1MB uses 8-bit weights; other results use
    bfloat16 (BF16) on one NVIDIA B200. nDCG@10 is normalized discounted cumulative gain at
    rank ten ($\times100$). Ruri: Ruri~v3; BGE: bge-reranker-v2-m3. Calibration
    uses development data.}
    \label{tab:long}
    \begin{tabular}{@{}llrl@{}}
    \toprule
    Benchmark or test & Metric & DF-Long & Comparison or scope \\
    \midrule
    S1MB English & Task Avg & \textbf{63.55} & DF-Large: 62.31 \\
    S1MB Japanese & Task Avg & \textbf{63.57} & Best public snapshot result: 60.05 \\
    \midrule
    JQaRA & nDCG@10 & \textbf{90.1} & Ruri: 86.9; BGE: 67.8 \\
    MIRACL Japanese & nDCG@10 & \textbf{93.7} & Ruri: 91.8; BGE: 90.1 \\
    MLDR Japanese & nDCG@10 & \textbf{98.9} & Ruri: 97.7; BGE: 97.6 \\
    \midrule
    Embedded instructions & Ignored (\%) & 99.9 & 899 of 900 trials; three languages \\
    Development decisions & ECE & 0.024 & 600 decisions \\
    \bottomrule
    \end{tabular}
\end{table}

\paragraph{Long-context serving and ranking.}
DF-Long extends the same decision interface to passage retrieval and long-context evidence processing (Table~\ref{tab:long}). Despite activating substantially fewer parameters, it slightly outperforms DF-Large on English decision tasks. It also achieves higher passage-ranking scores than both retrieval baselines across all three Japanese benchmarks. In the robustness evaluation, DF-Long successfully ignores 899 out of 900 embedded instructions across three languages. Its expected calibration error (ECE) is 0.024 on the development decisions used for temperature fitting. DF-Long also handles lengthy inputs efficiently, answering a question about a 27K-token document in just 0.41~s on the evaluated GPU.

\paragraph{Serving efficiency.}
The runner delivers its largest speedups when processing multiple decisions (Table~\ref{tab:runtime}). For eight questions sharing the same state, latency decreases from 71.7~ms to 29.4~ms, yielding a $2.44\times$ speedup. Even when processing eight independent states, the runner achieves a $2.23\times$ speedup. With four concurrent states and eight questions per state, throughput reaches 570 decisions per second.

\begin{figure}[tb]
  \centering
  \includegraphics[width=\linewidth]{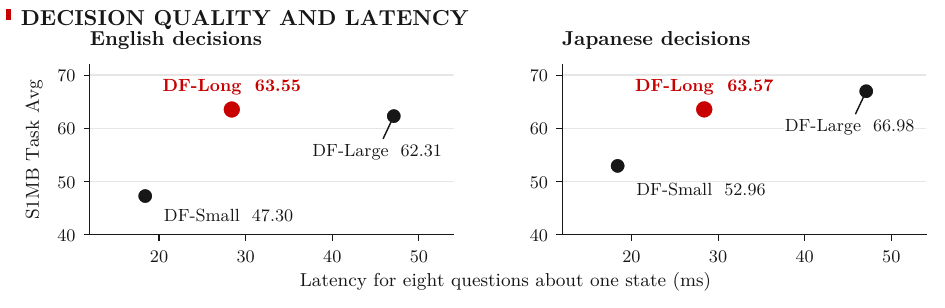}
  \caption{S1MB Task Avg versus eight-question latency. Quality scores use
  8-bit model packages; latency uses BF16 and the decision runner on one NVIDIA B200. Timings are separate from Table~\ref{tab:runtime}.}
  \label{fig:tradeoff}
\end{figure}

\begin{figure}[tb]
  \centering
  \includegraphics[width=\linewidth]{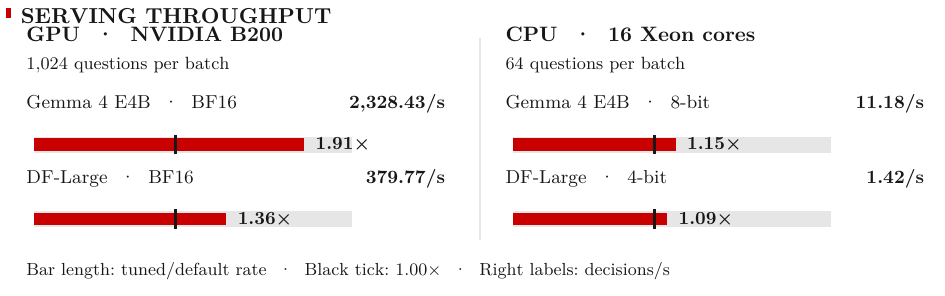}
  \caption{Serving throughput after tuning concurrency and the
  batched-token budget. Each bar compares tuned throughput with its own
  default setting; labels give the measured decisions per second.}
  \label{fig:inference}
\end{figure}

DF-Long achieves a favorable balance between decision quality and inference latency (Figure~\ref{fig:tradeoff}). Its latency approaches that of DF-Small while delivering substantially higher scores on both Japanese and English tasks. In a separate cross-model measurement, DF-Long processes an eight-question batch in approximately 28~ms, which is about 40\% faster than DF-Large. Further tuning of concurrency and the batched-token budget increases bulk throughput to approximately 2,328 decisions per second for the general-recipe E4B model and 380 decisions per second for DF-Large on a single GPU (Figure~\ref{fig:inference}). On 16 CPU cores, the E4B model achieves 11.2 decisions per second.

Importantly, these runtime optimizations preserve the decision readout. Using identical DF-Long BF16 weights, the runner and standard inference path differ by at most 0.2 points in their end-to-end S1MB English, S1MB Japanese, and JQaRA scores. The corresponding S1MB results using separate 8-bit model packages are reported in Table~\ref{tab:long}.

\begin{table}[tb]
    \centering
    \setlength{\tabcolsep}{7pt}
    \caption{Matched serving comparison for DF-Long in BF16 on one NVIDIA B200.
    The general-purpose path uses full-vocabulary readout and one request per
    question. Speedup is the latency ratio, or the throughput ratio for the
    concurrent workload.}
    \label{tab:runtime}
    \begin{tabular}{@{}p{6.0cm}rrr@{}}
    \toprule
    Workload & General path & Decision runner & Speedup \\
    \midrule
    Eight questions about one state & 71.7 ms & \textbf{29.4 ms} & \textbf{2.44$\times$} \\
    Eight states, one question each & 68.8 ms & 30.9 ms & 2.23$\times$ \\
    Four states, eight questions each & 364 decisions/s & \textbf{570 decisions/s} & 1.57$\times$ \\
    One question & 16.3 ms & 15.0 ms & 1.09$\times$ \\
    \bottomrule
    \end{tabular}
\end{table}

\subsection{Discussion}

These results highlight the importance of task-specific specialization alongside backbone scale. Despite activating only about 4 billion parameters per token, DF-Long outperforms its unspecialized 31 billion parameter backbone on Japanese decision tasks. Similarly, the compact distilled models achieve improvements of more than 20 points without increasing their backbone size. By incorporating long-context and relevance-judgment data, DF-Long further extends the range of supported applications while retaining the same decision interface and serving infrastructure.

The decision runner makes these specialized models practical to deploy. Shared-prefix execution is particularly effective when multiple questions refer to the same evidence, while option-only output reduces computation even for independent requests. 

\section{Related Work}

Model factories aim to streamline model development through reusable components and versioned experiments \citep{rowell2025factory}. Jev \citep{almeida2026jev} introduces a decision interface based on probabilities assigned to predefined answers. \citet{li2026} investigates structured answer-suffix scoring, while \citet{geuter2026} combines option-token logits with low-rank fine-tuning to develop open decision models. For Japanese decision tasks, \citet{tateno2026saihai} presents a specialized model and its benchmark evaluation. Building on these directions, the Decision Model Factory provides a unified workflow for the reproducible specialization, verification, and efficient deployment of models that share a common probabilistic decision interface.

\section{Conclusion and Outlook}

We presented the Decision Model Factory, a framework for building and deploying specialized, operator-controlled decision models as application requirements evolve. Through five DF models, we demonstrated that a single workflow can support Japanese-focused, distilled, and long-context specializations without requiring separate training or serving systems. The improvements in decision quality, together with the efficiency gains from the dedicated runner, demonstrate the practical value of combining model specialization with decision-specific inference. The recipe-based workflow also provides a path toward supporting additional tasks and languages while keeping training data, model weights, and deployment under the operator's control.

A promising direction for future work is to integrate PHOTON \citep{ichikawa2026photon} into the Decision Model Factory. PHOTON's hierarchical latent-stream architecture is designed to support efficient long-context and multi-query processing, making it particularly relevant to applications that require multiple decisions over shared documents. Incorporating PHOTON into the existing specialization, verification, and serving workflow could enable more memory-efficient processing of large numbers of parallel judgments. We leave this integration for future work.

\bibliographystyle{unsrtnat}
\bibliography{references}

\clearpage
\appendix

\section{Main Hyperparameters}
\label{app:settings}

For all models, we train LoRA adapters on the attention and feed-forward projections for one epoch, including the fused expert weights of Gemma~4~26B-A4B. All training runs use a 5\% warm-up fraction. The \emph{Score} loss weight is fixed at $\lambda=1$, and the candidate adapter strengths are $\Sigma=\{0.5,0.75,1\}$. Table~\ref{tab:hparams} summarizes the hyperparameters that vary across builds. For DF-Long, paired values correspond to the initial and long-input training stages, respectively. Effective batch size denotes the number of training decisions per optimizer step.

For the evaluated Gemma~4 backbones, the output transformation described in Section~\ref{sec:readout} uses logit soft-capping:
$z_k(x)=c\tanh(\nicefrac{w_k^{\top}h(x)}{c})$ with $c=30$. The runner applies the same transformation to the selected option logits. Of 18 DF-Long output-projection comparisons in BF16 precision, 17 produce bit-identical logits. The remaining comparison differs by only one BF16 rounding step, with probability differences below $2\times10^{-8}$. For Gemma~4~E4B, the final 18 layers reuse attention states computed in earlier layers, enabling the prompt-position optimization described in Section~\ref{sec:runtime}.

\begin{table}[tb]
    \centering
    \setlength{\tabcolsep}{4pt}
    \caption{Training settings for the DF models. A dash indicates that no
    teacher is used.}
    \label{tab:hparams}
    \begin{tabular}{@{}lrrrrr@{}}
    \toprule
    & DF-JA-Large & DF-Large & DF-Long & DF-Small & DF-JA-Small \\
    \midrule
    Adapter rank $r$ & 16 & 32 & 32 & 16 & 16 \\
    Learning rate & $10^{-4}$ & $10^{-4}$ & $10^{-4}$ / $5\times10^{-5}$ & $10^{-4}$ & $10^{-4}$ \\
    Effective batch size & 32 & 32 & 32 / 16 & 16 & 16 \\
    Maximum prompt (tokens) & 8,192 & 8,192 & 8,192 / 32,768 & 8,192 & 8,192 \\
    Teacher weight $\beta$ & -- & -- & -- & 0.5 & 0.5 \\
    \bottomrule
    \end{tabular}
    \end{table}
\end{document}